\documentclass[runningheads]{llncs}

\usepackage{graphicx}
\usepackage{booktabs}
\usepackage{graphicx}
\usepackage{pgfplots}
\pgfplotsset{width=\textwidth,compat=1.17}

\begin{document}

\title{CSEPrompts: A Benchmark of Introductory Computer Science Prompts}

\titlerunning{CSEPrompts - A Benchmark of Introductory Computer Science Prompts}

\author{Nishat Raihan\inst{1},
Dhiman Goswami\inst{1},
Sadiya Sayara Chowdhury Puspo\inst{1} \\
Christian Newman\inst{2},
Tharindu Ranasinghe\inst{3},
Marcos Zampieri\inst{1}}

\authorrunning{Raihan et al.}

\institute{George Mason University, Fairfax, VA, USA\\
\and
Rochester Institute of Technology, Rochester, NY, USA\\
\and
Aston University, Birmingham, UK\\
\email{mraihan2@gmu.edu} \\
}

\maketitle              

\begin{abstract}
Recent advances in AI, machine learning, and NLP have led to the development of a new generation of Large Language Models (LLMs) that are trained on massive amounts of data and often have trillions of parameters. Commercial applications (e.g., ChatGPT) have made this technology available to the general public, thus making it possible to use LLMs to produce high-quality texts for academic and professional purposes. Schools and universities are aware of the increasing use of AI-generated content by students and they have been researching the impact of this new technology and its potential misuse. Educational programs in Computer Science (CS) and related fields are particularly affected because LLMs are also capable of generating programming code in various programming languages. To help understand the potential impact of publicly available LLMs in CS education, we introduce CSEPrompts\footnote{\url{https://github.com/mraihan-gmu/CSEPrompts}}, a framework with hundreds of programming exercise prompts and multiple-choice questions retrieved from introductory CS and programming courses. We also provide experimental results on CSEPrompts to evaluate the performance of several LLMs with respect to generating Python code and answering basic computer science and programming questions.

\end{abstract}

\section{Introduction}
\label{sec:intro}

In the last decade, NLP models have evolved from n-gram and word embedding models (e.g., word2vec \cite{mikolov2013distributed}, GloVe \cite{pennington2014glove}) to advanced context-aware models like ELMo \cite{peters-etal-2018-deep} and BERT \cite{devlin2018bert}, significantly improving performance in various tasks \cite{rogers2020primer}. Recent developments in Large Language Models (LLMs), notably GPT-3 and GPT-4, have further revolutionized NLP by enabling human-like text generation and application in fields such as healthcare \cite{nori2023capabilities}, education \cite{tack2022ai} and many other novel tasks \cite{raihan2023sentmix,goswami2023offmix,raihan2023offensive}, marking a new era in generative AI.

Several recent studies have addressed the impact of GPT models on education \cite{lo2023impact,sok2023chatgpt,halaweh2023chatgpt}. While these models bring several opportunities in educational technology, such as enhanced writing assistants, intelligent tutoring systems, and automatic assessment tools, concerns arise from the misuse of technology, particularly in coding tasks. The study conducted by Savelka et al. \cite{savelka2023can} shows that while GPT scores may not meet course completion criteria, the model shows notable capabilities, including correcting solutions based on auto-grader feedback. Students may take advantage of this technology to generate complete essays and programming assignments obtaining artificially high grades. Furthermore, it has been shown that ChatGPT excels in debugging, bug prediction, and explanation but has reasoning and integration limitations \cite{surameery2023use}. Recent studies investigate the use of GPT models on assessment in domains such law \cite{katz2023gpt}, mathematics and computer science \cite{zhang2023exploring}, and medicine \cite{haruna2023gpt} evidencing the high quality of the models' output which would ``pass the bar exam'' \cite{katz2023gpt}. 

In this paper, we investigate the impact of LLMs on CS education and assessment by carrying out an evaluation of the performance in introductory CS and programming course assignments. While all aforementioned studies are restricted to GPT \cite{katz2023gpt,zhang2023exploring,haruna2023gpt}, we present a more comprehensive evaluation of models that goes beyond GPT. With the goal of enabling reproducibility, we create CSEPrompts, a framework comprising 219 programming prompts and 50 multiple-choice questions (MCQs) collected from coding websites and massive open online courses (MOOCs). We investigate the performance of eight models capable of generating both English text and programming code in Python.  We address the following research questions:

\begin{itemize}
\item RQ1: How well do state-of-the-art LLMs perform on introductory CS assignments compared to existing Benchmarks? 

\item RQ2: Is there a significant difference in the performance of LLMs when completing assignments from coding websites compared to academic MOOCs?

\item RQ3: Are state-of-the-art LLMs better at generating code or answering MCQs? 

\item RQ4: Are Code LLMs better at generating code and/or answering MCQs than raw LLMs? 

\end{itemize}

\section{Related Work}
Prior to the revolution of generative pre-trained models, most coding tasks involved tasks such as code completion, code infilling, comment generation, and similar tasks that are often handled employing BERT-like \cite{devlin2018bert} encoder-only models. Such models include CodeBERT \cite{feng2020codebert}, GraphCodeBERT \cite{guo2020graphcodebert} among others that are pre-trained on text-code pairs, often including ASTs (Abstract Syntax Trees) and CFGs (Control Flow Graphs) as well in the training corpus. However, encoder-only models are not primarily designed to be used as generative models and show subpar performance at Code Generation Tasks \cite{wang2021syncobert}. 

With the advent of generative models that are based on either encoder-decoder \cite{wang2021codet5} or decoder-only \cite{roziere2023code} architecture, they start to show better code-generation capabilities - evidenced by the survey conducted by Zan et al. \cite{nl2code}. Hence, the need for a unified benchmark arises. Such datasets include HumanEval (introduced with GPT3.5 \cite{OpenAI2023GPT4TR}) and MBPP \cite{austin2021program} that contain a set of coding prompts, paired with human-generated solutions and three test cases for each task. Other similar datasets include CONCODE \cite{iyer2018mapping}, DS-1000 \cite{lai2023ds} and the extensions of the two previously mentioned datasets as HumanEval+ \cite{liu2023your} and MBPP+ \cite{guo2023instruction}. Another similar task 

However, these datasets do not focus on coding tasks that are often used in the educational domain - rather mostly common tasks that occur in software development or other aspects. The coding tasks in the education domain, focus more on the in-depth knowledge of the specific programming language that evaluates the core understanding of the syntax and semantics of the language, making them significantly different from the prompts that are included in the existing benchmarks. To bridge this gap, we introduce CSEPrompts, a framework with hundreds of programming exercise prompts and multiple-choice questions retrieved from introductory CS and programming courses. Each prompt is paired with five test cases, compared to three for most benchmarks.

\section{CSEPrompts}

We introduce CSEPrompts, a novel evaluation framework consisting of coding prompts from coding websites and academic MOOCs (Table \ref{tab:link}). CSEprompts features a total of 269 exercise prompts as shown in Table \ref{tab:tableDS}. We include links to the questions and answers where the interested reader can download the data. 

\begin{table*}[h]
    \caption{List of Coding Websites \& MOOCs}
    \label{tab:link}
    \centering
    \scalebox{0.85}{
        \begin{tabular}{ll}
            \toprule
            \textbf{Name} & \textbf{Link}\\
            \midrule
            CodingBat & \url{https://codingbat.com/python}\\
            Learn Python & \url{https://www.learnpython.org}\\
            Edabit & \url{https://edabit.com/challenges/python3}\\
            Python Principles & \url{https://pythonprinciples.com/challenges/}\\
            Hacker Rank & \url{https://www.hackerrank.com/domains/python}\\
            Edx & \url{https://www.edx.org}\\
            Coursera & \url{https://www.coursera.org}\\
            CS50 (Harvard) & \url{https://learning.edx.org/course/course-v1:HarvardX+CS50S+Scratch/home}\\
            PforE (UMich) & \url{https://www.coursera.org/learn/python/home}\\
            CS1301xI (GT) & \url{https://learning.edx.org/course/course-v1:GTx+CS1301xI+1T2023/home}\\
            CS1301xII (GT) & \url{https://learning.edx.org/course/course-v1:GTx+CS1301xII+1T2023/home}\\
            CS1301xIII (GT) & \url{https://learning.edx.org/course/course-v1:GTx+CS1301xIII+1T2023/home}\\
            CS1301xIV (GT) & \url{https://learning.edx.org/course/course-v1:GTx+CS1301xIV+1T2023/home}\\
            \bottomrule
        \end{tabular}
    } 
\end{table*}

\paragraph{\textbf{Coding Websites}} We choose five leading online resources for introductory Python learning, as detailed in Table \ref{tab:tableDS}. These include \textit{CodingBat}, offering a variety of coding challenges, \textit{LearnPython} with its interactive tutorials, \textit{Edabit} for a wide range of programming problems, \textit{Python Principles} emphasizing practical exercises, and \textit{HackerRank}, known for its diverse coding challenges and competitions. Our selection criteria for these platforms center on their interactivity, diversity of challenges, and structured learning approaches. Interactive methods, as provided by \textit{LearnPython} and \textit{Python Principles}, enable users to engage actively in learning. Meanwhile, \textit{Edabit} and \textit{HackerRank} cater to a broad spectrum of skill levels. These platforms are bolstered by strong communities and offer instant feedback, crucial for effective learning in programming.

\paragraph{\textbf{MOOCs}} Our study includes programming assignments, tasks, and multiple-choice questions from six MOOCs offered by Harvard University, the University of Michigan, and the Georgia Institute of Technology. These courses are available on online platforms like edx and Coursera, focusing on introductory Python programming for both beginners and those with some programming background. Details of these courses are presented in Table \ref{tab:tableDS}. \textit{Harvard's CS50}, on edx, introduces basic programming concepts. \textit{Michigan's Programming for Everybody (PforE)}, found on both edx and Coursera, is designed for programming beginners. Georgia Tech's courses, \textit{CS1301xI}, \textit{CS1301xII}, \textit{CS1301xIII}, and \textit{CS1301xIV}, cover a range of Python topics from beginner to advanced levels. These courses include practical programming exercises, but we exclude tasks involving File I/O or Command-Prompt/Terminal due to the limitations of Language Learning Models (LLMs) in such interactions. Finally, along with the coding tasks, we also gather several MCQs each containing a set of 5 to 10 options containing the correct answer and multiple distractors.  We gather a total of fifty such MCQs from four courses from GT, mentioned in Section - CS1301xI, CS1301xII, CS1301xIII and CS1301xIV. A brief description is provided in the Table \ref{tab:tableDS}.

\begin{table*}[!h]
\caption{Summary of Coding Prompts from Various Sources}
\centering
\resizebox{\textwidth}{!}{
\begin{tabular}{llc|llc|llc}
\hline
\multicolumn{3}{c|}{\textbf{Coding Websites}} & \multicolumn{3}{|c|}{\textbf{MOOCs - Coding Prompts}} & \multicolumn{3}{|c}{\textbf{MOOCs - MCQs}} \\
\hline
& Platform & Prompts & University & Course & Prompts & University & Course & Prompts \\
\hline
& CodingBat & 24 & Harvard & CS50 & 29 & GT & CS1301xI & 20 \\
& LearnPython & 16 & UMich & PforE & 7 & GT & CS1301xII & 8 \\
& Edabit & 29 & GT & CS1301xI & 11 & GT & CS1301xIII & 6 \\
& Python Principles & 26 & GT & CS1301xII & 20 & GT & CS1301xIV & 16 \\
& HackerRank & 23 & GT & CS1301xIII & 17 & & & \\
\hline
\multicolumn{2}{r}{\textbf{Total}} & \textbf{118} & \multicolumn{2}{|r}{\textbf{Total}} & \textbf{101} & \multicolumn{2}{|r}{\textbf{Total}} & \textbf{50} \\
\hline
\end{tabular}
}

\label{tab:tableDS}
\end{table*}

\paragraph{\textbf{Dataset Statistics}} The prompts are mostly shorter in the coding sites comparatively. A few key statistics are presented in Table \ref{tab:prompts_statistics}, about the prompts. We gather at least 5 test cases for each prompt for our experiment. In most cases, the test cases are taken from the platforms from where the prompts themselves are taken. In cases where we have less, we generate more test cases using pynguin \cite{lukasczyk2022pynguin}, an open-source unit test case generator tool for Python. For the MCQs, we gather the correct answer(s) from the platforms themselves. We further gather responses generated by the LLMs for each prompt. The responses are cleaned manually and only the code snippets are kept. We label each code snippet based on how many test cases they passed.

\begin{table}[ht]
\caption{Statistics for Prompts}
\centering
\begin{tabular}{l|c|c|c}
\hline
\textbf{Metric} & \textbf{CodingSites} & \textbf{Academic} & \textbf{MCQ} \\
\hline
Total Prompts & 118 & 101 & 50 \\
Max. No. of Tokens & 101 & 372 & 221 \\
Min. No. of Tokens & 5 & 17 & 15 \\
Mean No. of Tokens & 28 & 158 & 106 \\
Standard Deviation & 16 & 72 & 51 \\
\hline
\end{tabular}

\label{tab:prompts_statistics}
\end{table}

\paragraph{\textbf{Data Collection Strategy}} Unlike HumanEval \cite{OpenAI2023GPT4TR} or MBPP \cite{austin2021program} benchmarks that are generated by human, specifically for the purpose of evaluating code generation capabilities of Large Language Models; we take a different approach. In order to obtain the academic nuances and intricacies of coding tasks, we collect them from real-world academic courses and coding websites. They are collected manually without any web scrapper or extractor. It is also ensured manually that no prompts and/or tasks are duplicated.

\section{Experiments}
\paragraph{\textbf{LLMs}}
There are several proprietary LLMs like GPT-3.5 \cite{OpenAI2023GPT4TR}, PaLM-2 \cite{anil2023palm} etc., and Open-Source LLMs like Llama-2 \cite{touvron2023llama}, Falcon \cite{penedo2023refinedweb}, StableLM \cite{stabilityai_stablelm} etc. that are used for a wide variety of tasks, including coding problems as well. The eight LLMs used for our experiments are briefly mentioned in Table \ref{tab:llms}. For the MCQs, the prompt has a minor modification as shown in Figure \ref{fig:prompt2}.

\begin{table*}[h]
\caption{LLMs Used on CSEPrompts}
    \begin{tabular}{p{3.5cm}|p{3.5cm}|p{2.75cm}|p{2cm}}
    \hline
    \textbf{LLM} & \textbf{Parameter} & \textbf{Model Type} & \bf Reference \\
    \hline
    GPT3.5 & 175B & Base & \cite{OpenAI2023GPT4TR} \\
    Llama2 & 7B & Base & \cite{touvron2023llama} \\
    Falcon & 7B & Base & \cite{penedo2023refinedweb} \\
    MPT & 7B & Base & \cite{mosaicml_mpt30b} \\
    Code-Llama & 7B & Fine-tuned & \cite{roziere2023code} \\
    StarCoder & 7B & Fine-tuned & \cite{li2023starcoder} \\
    WizardCoder & 7B & Fine-tuned & \cite{luo2023wizardcoder} \\
    Mistral & 7B & Base & \cite{jiang2023mistral} \\
    \hline
    \end{tabular}
    
    \label{tab:llms}
\end{table*}

\paragraph{\textbf{Code Generation}}
We first prepare our prompts and then test each model for all the tasks and questions from CSEPrompt. The prompts have a simple format, as shown in Figure \ref{fig:prompt1} and \ref{fig:prompt2}. Each model is prompted with the same prompt. The models generate code, but they also generate more texts, including pseudo-codes, explanations, etc. The responses are then cleaned manually to exclude everything other than the code. The codes are then tested for all the test cases using the pytest \cite{pytest_2023} framework, which facilitates the easy creation of small and readable unit tests for Python codes. 

\begin{figure}[!h]
\centering
\scalebox{.9}{
\begin{tikzpicture}[node distance=1.35cm]
    \tikzstyle{block} = [rectangle, draw, fill=red!20, text width=\linewidth, text centered, rounded corners, minimum height=4em]
    \tikzstyle{operation} = [text centered, minimum height=1em]
    \node [block] (rect1) {\textit{'You are a helpful AI assistant. You are given the following problem: '}};
    \node [block, below of=rect1] (rect2) {\textbf{Write a function named capital\_indexes. The function takes a single parameter, which is a string. Your function should return a list of all the indexes in the string that have capital letters.}};
    \node [block, below of=rect2] (rect3) {\textit{'Please write a Python code snippet to solve the problem. Thanks.'}};
\end{tikzpicture}
}
\caption{Sample Prompt for Coding Tasks.}
 \label{tab:prompt1}
\label{fig:prompt1}
\end{figure}
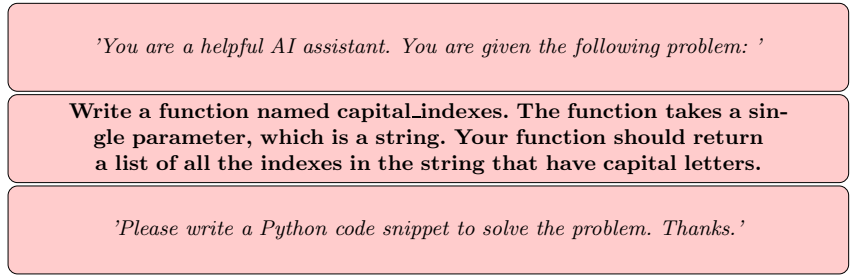

\begin{figure}[!h]
\centering
\scalebox{.9}{
\begin{tikzpicture}[node distance=1.5cm]
    \tikzstyle{block} = [rectangle, draw, fill=blue!20, text width=\linewidth, text centered, rounded corners, minimum height=1em]
    \tikzstyle{operation} = [text centered, minimum height=1em]

    \node [block] (rect3) {\textit{'You are a helpful AI assistant. You are given a Multiple Choice Question. '}};
    \node [block, below of=rect3] (rect4) { \textbf{ (False and True) or (False or True) \newline Is this statement resolved to True or False? 
    \begin{itemize}
        \item True
        \item False
        \item Statement will not compile
    \end{itemize}}};
    \node [block, below of=rect4, yshift=0mm] (rect5) {\textit{'You need to pick one or multiple correct answers from the given ones. Thanks.'}};
\end{tikzpicture}
}
\caption{Sample Prompt for MCQs.}
\label{fig:prompt2}
\end{figure}
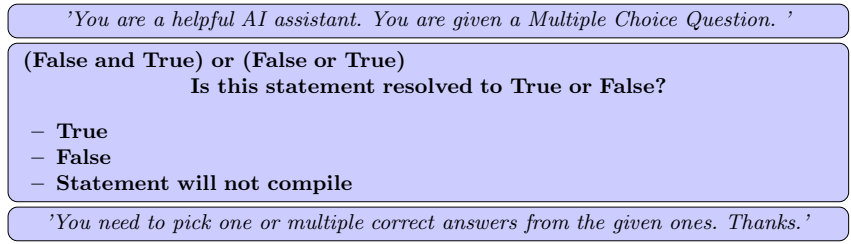

\section{Results}

For the Code Generation task, there are a few widely used benchmark datasets like HumanEval \cite{OpenAI2023GPT4TR} and MBPP \cite{austin2021program}. HumanEval has 164 coding tasks paired with three test cases each and MBPP compiles a larger datset with 974 tasks, also with 3 test cases each. Code LLMs often present their results on these datasets based on the \textit{Pass@K} metric - which means how many attempts the model take to pass all the test cases. We compare CSEPrompts with these two benchmark datasets based on Pass@1. Since these are Code Generation tasks, we exclude the MCQ tasks from CSEPrompts during the comparison (shown in Figure 3).

\begin{figure}[!h]
\centering
\scalebox{0.75}{
\begin{tikzpicture}
\begin{axis}[
    ybar,
    bar width=.25cm,
    width=\textwidth,
    height=.5\textwidth,
    enlarge x limits=0.15,
    legend style={at={(0.5,1.15)},
      anchor=north,legend columns=-1},
    ylabel={Pass@1},
    symbolic x coords={GPT3.5,Mistral,WizardCoder,StarCoder,CodeLLaMA,LLaMA,Falcon,MPT},
    xtick=data,
    x tick label style={rotate=45,anchor=east},
    ]
\addplot coordinates {(GPT3.5,73.2) (Mistral,28.7) (WizardCoder,48.2) (StarCoder,24.4) (CodeLLaMA,34.1) (LLaMA,24.1) (Falcon,15.7) (MPT,7.2)};
\addplot coordinates {(GPT3.5,81.7) (Mistral,50.1) (WizardCoder,56.6) (StarCoder,57.7) (CodeLLaMA,57.6) (LLaMA,41.2) (Falcon,19.3) (MPT,12.4)};
\addplot coordinates {(GPT3.5,83.1) (Mistral,74.6) (WizardCoder,48.3) (StarCoder,66.1) (CodeLLaMA,30.5) (LLaMA,44.9) (Falcon,23.7) (MPT,11.9)};
\addplot coordinates {(GPT3.5,71.3) (Mistral,22.8) (WizardCoder,28.7) (StarCoder,34.7) (CodeLLaMA,39.6) (LLaMA,13.9) (Falcon,0.9) (MPT,3.9)};
\legend{HumanEval,MBPP,CSEPrompts-MOOCs,CSEPrompts-Academic}
\end{axis}
\end{tikzpicture}
} 
\caption{Comparing CSEPrompts with HumanEval and MBPP based on \textbf{Pass@1}.}
\end{figure}
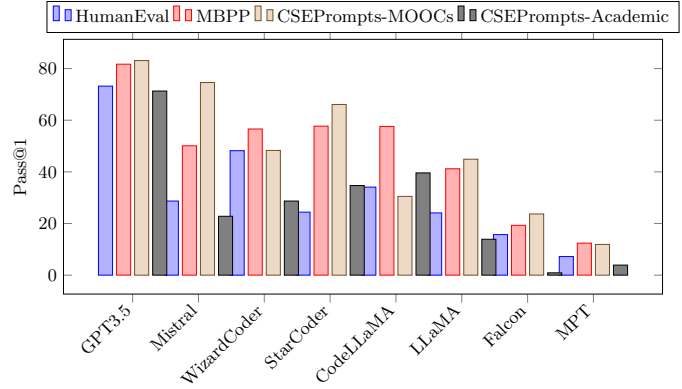

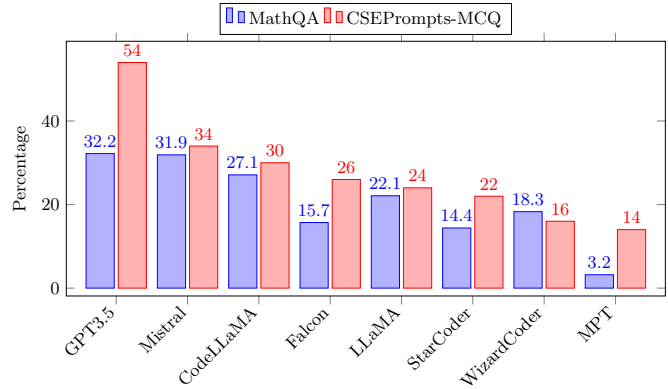
\begin{figure}[!h]
\label{fig:enc-perform}
\centering
\scalebox{0.75}{
\begin{tikzpicture}

\begin{axis}[
    ybar,
    bar width=.5cm,
    width=\textwidth,
    height=.5\textwidth,
    legend style={at={(0.5,1.15)},
      anchor=north,legend columns=-1},
    ylabel={Percentage},
    symbolic x coords={GPT3.5,Mistral,CodeLLaMA,Falcon,LLaMA,StarCoder,WizardCoder,MPT},
    xtick=data,
    x tick label style={rotate=45,anchor=east},
    nodes near coords,
    nodes near coords align={vertical},
    ]
\addplot coordinates {(GPT3.5,32.2) (Mistral,31.9) (CodeLLaMA,27.1) (Falcon,15.7) (LLaMA,22.1) (StarCoder,14.4) (WizardCoder,18.3) (MPT,3.2)};
\addplot coordinates {(GPT3.5,54) (Mistral,34) (CodeLLaMA,30) (Falcon,26) (LLaMA,24) (StarCoder,22) (WizardCoder,16) (MPT,14)};
\legend{MathQA,CSEPrompts-MCQ}
\end{axis}
\end{tikzpicture}
}
\caption{Comparing CSEPrompts-MCQ with MathQA based on Zero Shot Prompting (in percentage).}
\end{figure}

For the MCQ task, however, there are no similar benchmarks to compare with. The most similar existing dataset is the MathQA-Python benchmark (introduced with MBPP \cite{austin2021program}). This contains coding-related question-answer pairs in Python. However, no work explores the Code LLMs' expertise in the aspect of Multiple Choice Question Answering. In contrast, the Academic-MCQ subset of CSEPrompts is the first dataset Code-MCQ benchmark in the domain. We compare the results for several LLMs on it and MathQA-Python. As shown in Figure 4, the MCQ task is comparatively easier for the models compared to QA tasks - most likely because more context and a set of possible answers guide the model to generate a better answer.

\section{Conclusion and Future Work}

In this paper, we evaluated the output of different LLMs when answering MCQs and completing programming assignments in Python compiled from introductory CS courses. We created CSEPrompts, 
an evaluation framework containing programming prompts and MCQs curated from different online coding websites, academic platforms, and programming courses. We evaluated the performance of eight state-of-the-art LLMs and showed their detailed performance along with an error analysis. We revisit the four RQs posed in the introduction. 

\paragraph{RQ1: How well do state-of-the-art LLMs perform on introductory CS assignments compared to existing Benchmarks?} All models tested produced high-quality output, thus performing well on CSEPrompts with GPT outperforming the other seven models. Compared to existing benchmarks, the LLMs perform better at \textit{CSEPrompts [MOOCs]} and worse on \textit{CSEPrompts [Academic]}.

\paragraph{RQ2: Is there a significant difference in the performance of LLMs when completing assignments from coding websites compared to academic MOOCs?} The prompts from coding websites proved to be easier for most of the LLMs than those from academic MOOCs. This suggests that the prompts from academic MOOCs are more challenging/advanced than those from coding websites. 
  
\paragraph{RQ3: Are state-of-the-art LLMs better at generating code or answering MCQs?} As LLMs are developed primarily to generate text, our assumption is that LLMs would perform better at generation text than code, however, the LLMs we evaluated, are better at generating code than answering the MCQs.
   
\paragraph{RQ4: Are Code LLMs better at generating code and/or answering MCQs than raw LLMs?} GPT3.5 being the bigger model among the ones we tested - outperformed others on all tasks - while being a general-purpose LLM. But for most cases, general-purpose LLMs perform better at MCQs and Code LLMs perform better on coding tasks.

In future work, we would like to carry out an even more comprehensive study by increasing the number of coding prompts and MCQs in the framework. Furthermore, we would like to incorporate other LLMs not included in this study.


\end{document}